\documentclass[conference]{IEEEtran}
\IEEEoverridecommandlockouts
\usepackage{cite}
\usepackage{amsmath,amssymb,amsfonts}
\usepackage{algorithmic}
\usepackage{graphicx}
\usepackage{textcomp}
\usepackage{xcolor}
\def\BibTeX{{\rm B\kern-.05em{\sc i\kern-.025em b}\kern-.08em
    T\kern-.1667em\lower.7ex\hbox{E}\kern-.125emX}}
\begin{document}

\title{Radar detection rate comparison through a mobile robot platform at the ZalaZONE proving ground}

\author{\IEEEauthorblockN{V\'ictor J. Exp\'osito Jim\'enez, Christian Schwarzl}
\IEEEauthorblockA{\textit{Dependable Systems Group, Department E} \\
\textit{Virtual Vehicle Research Center GmbH}\\
Graz, Austria \\
victor.expositojimenez@v2c2.at}
\and
\IEEEauthorblockN{Szil\'ard J\'osvai}
\IEEEauthorblockA{
\textit{Automotive Proving Ground Zala Ltd.}\\
Zalaegerszeg, Hungary \\
szilard.josvai@apz.hu}
}

\maketitle

\begin{abstract}
	
Since an automotive driving vehicle is controlled by Advanced Driver-Assistance Systems (ADAS) / Automated Driving (AD) functions, the selected sensors for the perception process become a key component of the system. Therefore, the necessity of ensuring precise data is crucial. But the correctness of the data is not the only part that has to be ensured, the limitations of the different technologies to accurately sense the reality must be checked for an error-free decision making according to the current scenario. In this context, this publication presents a comparison between two different automotive radars through our self-developed robot mobile platform called SPIDER, and how they can detect different kinds of objects in the tests carried out at the ZalaZONE proving ground.

\end{abstract}


\section{INTRODUCTION}
\label{section:introduction}

In recent years, responsible functions and systems for the Automated Driving~(AD) and Advanced Driver-Assistance Systems~(ADAS) of a vehicle have become highly complex, and different type of sensors has to be included in order to give the system the necessary redundancy and reliability to make the automated driving vehicles as safest as possible. In this context, standards to give the guidelines to reach this safety goal in the automotive domain already exist,~\cite{ISO21448}~\cite{ISO26262}~\cite{SAEJ3061}, and those try to cover all failure issues from sensor fails to technical limitations.   

Each technology has its advantages and disadvantages, which make it suitable for each situation. Only the combination of all technologies makes the system reliable in all circumstances. For example, lidar performance is reduced in harsh weather environments~\cite{7795565}, and most of the cameras are sensitive to low light environments. On the other hand, radar technology presents better performance against low-light and harsh environments, but its resolution to classify and detect some objects are behind the other technologies. Therefore, the data fusion of different types of sensors is compulsory in the automated driving domain. A state-of-the-art survey is shown by Wang et al. in~\cite{8943388} in which fusion strategies are explained and categorized. An especial mention was done in the application of deep learning in this process and further fusion techniques improvements.

In the scope of this publication, radar technology has been researched for years~\cite{1278800} in which the different available work frequencies are used. For example, the authors in~\cite{7060385} explain how they can reach a three centimeters precision in distance up to three meters using 24GHz frequency. In other research, the authors in~\cite{6875687} highlight the long-range of using 77GHz frequency and explain how to improve the pedestrian and tracking detection in this frequency range. Researchers are focused on the automotive domain along these years. Authors in~\cite{8126389} depict how the radars add significant value to automotive environment perception to enable fully automated driving due to their performance, reliability, and cost-efficiency. The publication shows technology strengths are still necessary to reach the high requirements to sensor robustness, however, it also points out its difficulty to interpret the obtained data from radars in comparison with other sensors technologies. 

The main goal of this publication is to give a brief comparison between the performance of two different radar technologies from the tests carried out on the proving ground. Section~\ref{section:spider} introduces the mobile robot platform in which the radar sensors have been integrated. In the next section, a short explanation of the proving ground, where the tests are done, is given. Then, the scenario definition and evaluation are explained in section~\ref{section:evaluation}. Finally, Section~\ref{section:conclusions} describes our conclusion according to the obtained results and the further steps to follow. 

\section{SMART PHYSICAL DEMONSTRATION AND EVALUATION ROBOT}
\label{section:spider}

\begin{figure}
	\centering
	\includegraphics[width=0.45\textwidth]{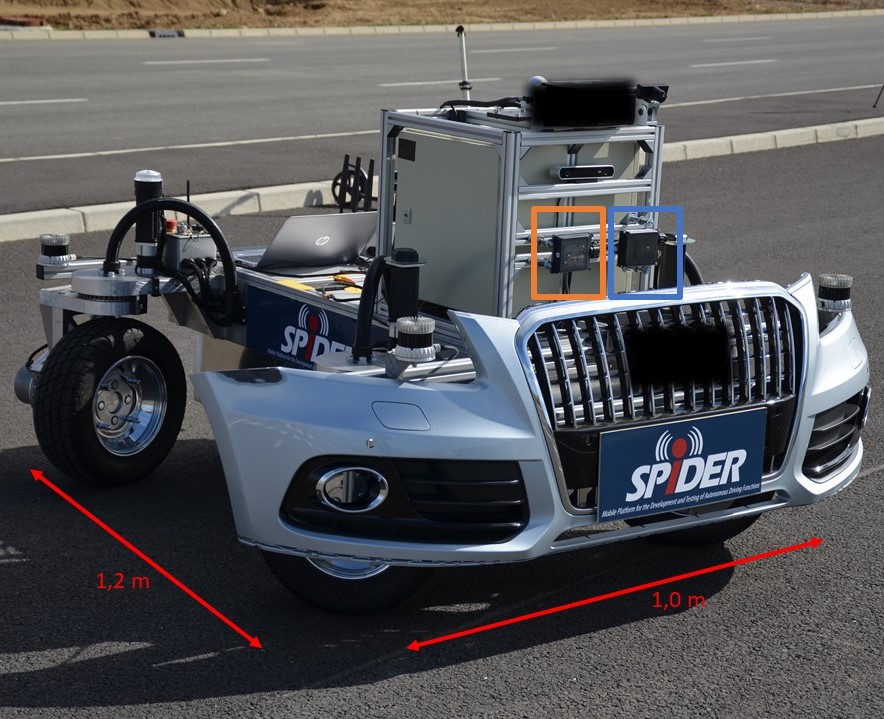}
	\caption{Smart PhysIcal Demonstration and Evaluation Robot (SPIDER). Orange Box: Long-Range Radar, Blue Box: Short Ranger Radar. [Notice that some sensors were covered up due to non-disclosure agreement] }
	\label{fig:spider}
\end{figure}

The radars are integrated into our \textit{“Smart PhysIcal Demonstration and Evaluation Robot” (SPIDER)}~\cite{spider} as shown in Figure~\ref{fig:spider}, which is a self-developed mobile platform for the development and evaluation of ADAS/AD functions and allowing for reproducible testing of perception systems, vehicle software and control algorithms under real world conditions. It is designed to be used in a wide variety of scenarios and adverse environmental conditions such as fog or light rain. Its high flexibility gives us the possibility of multiple and different modules and sensors can be mounted and unmounted to be used in different scenarios to analyze and compare the impact of the diverse configurations in different environments and weather conditions. Within the current configuration, it integrates high precision GNSS module, four lidars, 3D cameras, and the two evaluated radars among others.  

Additionally, the SPIDER allows developers high maneuverability to execute tests with no time losses due to the usage of four independent wheels which gives SPIDER the possibility of omnidirectional movements, including 360° or side-walks movements. SPIDER also include several safety functions, both software and hardware implemented, to avoid human harm, for example, the Collision Avoidance (CoA) function. 

\begin{figure}
	\centering
	\includegraphics[width=0.45\textwidth]{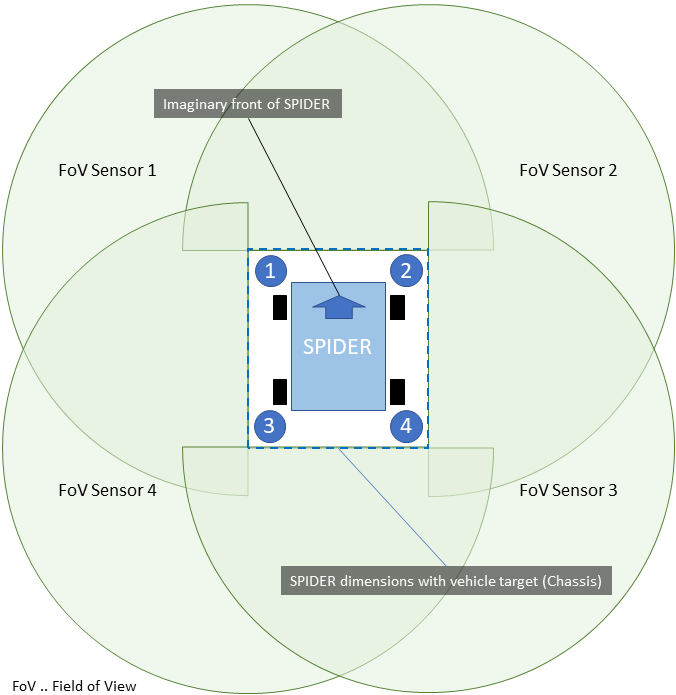}
	\caption{Lidars layout on the SPIDER}
	\label{fig:spider_coa}
\end{figure}

The CoA function is a safety-critical function and prevents the collision of the SPIDER with surrounding environment objects to avoid damage and most importantly human harm. The CoA uses several sensors, which constantly measure the distance to environment objects. Figure~\ref{fig:spider_coa} shows the distribution of the four lidars used to give the function the necessary sensor redundancy specially in the surrounding closest to the SPIDER. Although it is intended to be operated in a closed environment like a proving ground, where the access of humans is prohibited. However, to ensure maximum safety, the CoA shall detect humans (or objects) approaching the SPIDER from an arbitrary angle and reduce speed or initiate an emergency brake if they come too close. The platform also integrates a Path Tracking (PaT) function, which allows to load pre-defined paths to the SPIDER, which in combination with the integrated precise localization system, makes the SPIDER to autonomously reproduce the same path multiple times.  

The SPIDER Low-Level Control (LCC) is carried out by the Automotive Realtime Integrated NeXt Generation Architecture (AURIX)~\cite{aurix} platform. The responsible for the High-Level Control (HLC) is the Robotics Operating System (ROS)~\cite{ros}. ROS is an open-source middleware that allows building robot applications with great flexibility by using an extensive range of third-party libraries and tools for different domains such as Autoware.AI~\cite{autoware}, which is specially focused on the automated driving field. SPIDER integrates a High-Performance Computer (HPC) with a dedicated Graphical Process Unit (GPU), which allows executing high-demanding processes such as AI or data fusion.
 
\section{ZALAZONE PROVING GROUND}
\label{section:zalazone}

\begin{figure}
	\centering
	\includegraphics[width=0.47\textwidth]{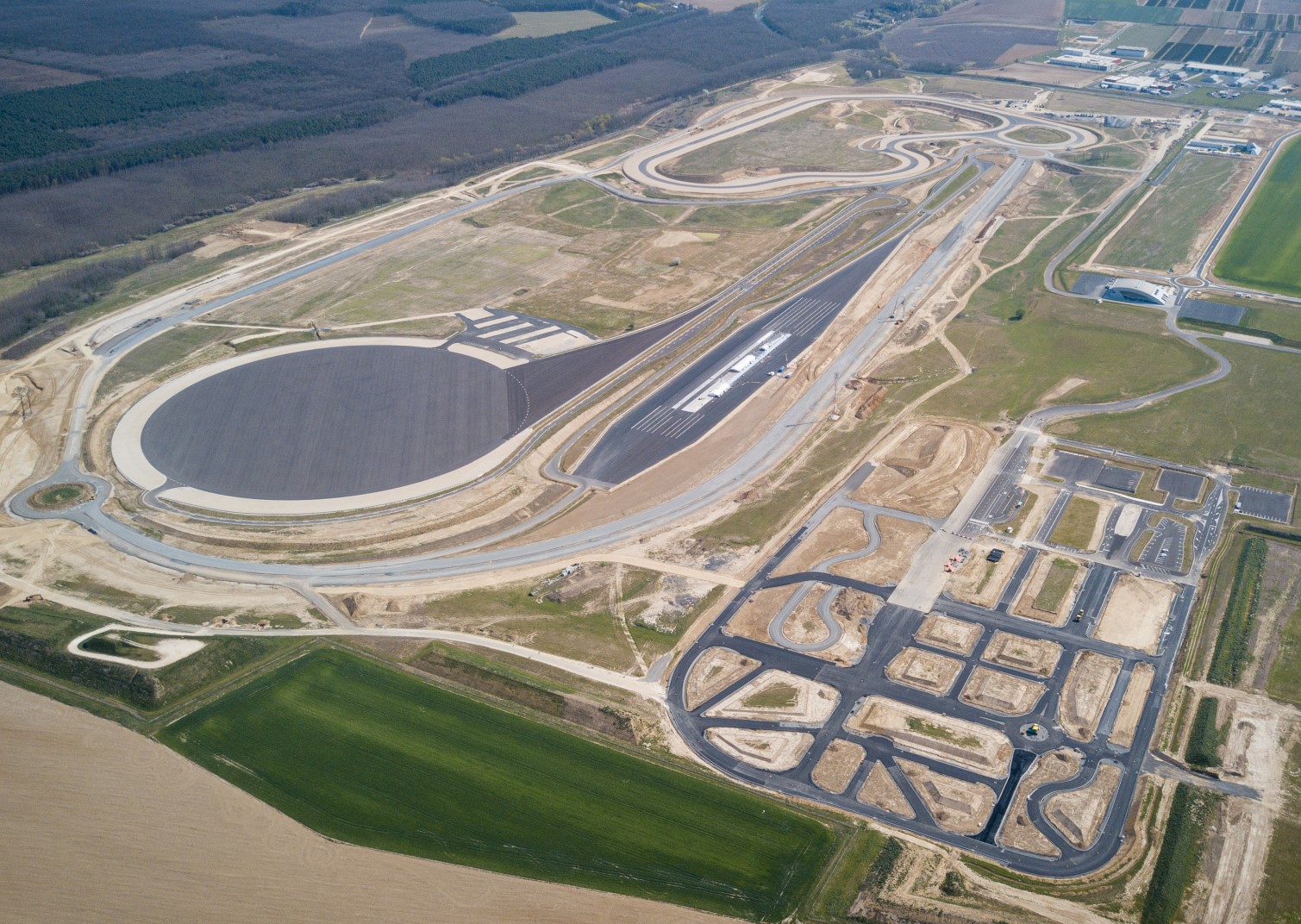}
	\caption{ZalaZONE proving ground aerial view }
	\label{fig:zalazone}
\end{figure}

The tests and measurements explained in this publication were performed at the ZalaZONE Proving Ground~\cite{szalay2019novel}. ZalaZONE - located at Zalaegerszeg, Hungary – has a unique design when it comes to the development of automated driving technologies. Its core function, being used not only for classical vehicle dynamics and endurance testing, but focusing on the special testing requirements of connected and automated vehicles. The main goal is to establish a full-range validation facility for future mobility and communication technologies. Figure~\ref{fig:zalazone} shows an aerial view of the whole complex formed by different sections for specific purposes. The 300m diameter vehicle dynamic plate, : the braking surface with 8 different types of braking lane, the 2 km long high-speed handling course and the smart city zone (Z-City) are already available for the testers, since May 2019. The remaining modules, including a motorway section, rural road section and the high speed oval are being constructed. The proving ground is planned to be finished by the end of 2021. Supporting the extensive testing of V2X capabilities ZalaZONE has already two independent 5G new radio communication technology implementations and will also be fully covered by ITS-G5 (DSRC) road-side units. Moreover, ZalaZONE is not only a proving ground. Its Research and Innovation Center acts as a catalytic hub in which many actors from the automated driving field can collaborate and interact to boost the current development of the domain~\cite{zalazone_validation_ecosystem}.

The SPIDER tests were carried out in the smart city zone (Z-City), which is an artificial metropolitan area, that was designed to provide a realistic environment for safe urban testing. In full implementation, Z-City will offer 5 separated urban zones that can be used parallel by testers, namely the low-speed parking area (i), the high-speed multi-lane area (ii), the downtown area (iii), the suburban area (iv) and the T-junction area (v). Each of these sections will have its characteristic lane distribution, special intersections, road segments, road furniture and building facades, fabricated from real constriction material. Although only phase one of Z-City is finished, it was enough to provide us the resources that were needed to execute the tests explained in this publication. 

\section{SCENARIO DEFINITION AND EVALUATION}
\label{section:evaluation}

\begin{figure*}
	\centering
	\includegraphics[width=1.00\textwidth]{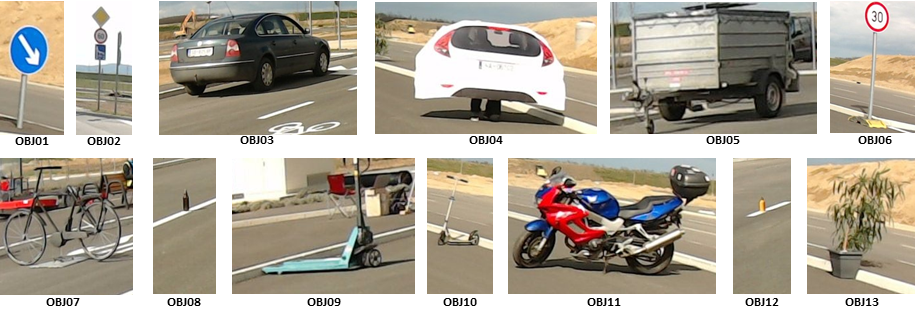}
	\caption{OBJ01: Traffic signal 01, OBJ02: Traffic signal 02, OBJ03: VW Passat car, OBJ04: Dummy car, OBJ05: Trailer, OBJ06: Traffic signal 03, OBJ07: Dummy bicycle, OBJ08: Empty glass bottle, OBJ09: Elevator, OBJ10: Scooter, OBJ11: Motorbike, OBJ12: Metal can, OBJ13: Flower pot}
	\label{fig:objects_list}
\end{figure*}

The main purpose of the work presented in this publication is to make a comparison between the detection performance of the two radar mounted on the SPIDER. Both radar models are a long-range radar (LRR), Continental Radar ARS408~\cite{ars408}, which works in the 77 GHz frequency range, and a short-range radar (SRR), Continental Radar SRR208~\cite{srr208}, which works in the 24 GHz. 

Three scenarios were defined for this evaluation. In each case, different objects have been placed along the scenario and it will be checked when the radars are able to detect the specific object. Figure~\ref{fig:objects_list} shows all used objects, which have different sizes and shapes in order to identify how these parameters can affect the detection. 

\begin{figure*}
	\centering
	\includegraphics[width=1.00\textwidth]{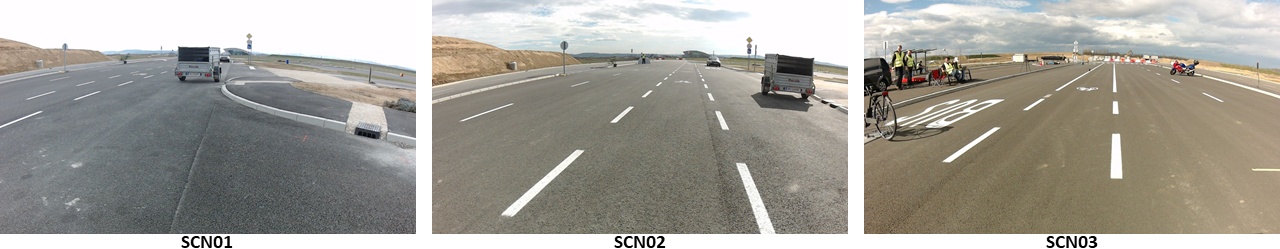}
	\caption{SCN01: Scenario 1, SCN02: Scenario 2, SCN03: Scenario 3}
	\label{fig:use_cases}
\end{figure*}

First (SCN01) and second scenarios (SCN02) are defined in the same area. This place is asphalted, flat, and straight. The area length is approximately seventy-five meters long.
The difference between the SCN01 and SCN02 is the place in which the SPIDER is located. 
In the SCN01, the robot is located in front of one of the right line of objects. On the other hand, the SPIDER is located in the middle of both lines of objects in the SCN02.
The third scenario (SCN03) is located in an area of similar characteristics than the previous scenarios but, in this case, different objects are tested. Figure~\ref{fig:use_cases} shows the vision of each scenario from the robot perspective. 

Table~\ref{table:radar_fov} shows the different Field of Views (FoV) that of each radar. 
According to this information, the common FoV for both radars is an azimuth of $\pm$40° and a range of 50 meters. Therefore, all objects from the defined scenarios are located in these area boundaries to ensure they are within radars' range.

The integration of the sensors was done in ROS and the data for each scenario was recorded for approximately one minute in which the the SPIDER stands still in place. To improve the quality of the data output, some filters~\cite{aintein_filters} were used, which soften the data output to avoid some jumps that may occur.

\setlength{\tabcolsep}{0.8em} 
{\renewcommand{\arraystretch}{1.2}
\begin{table}[h]
	\caption{Field of View of ARS408 and SRR208 radars}
	\label{table:radar_fov}
	\begin{center}
		\begin{tabular}{|c||c|c|}
			\hline
			Distance & Short-Range Radar & Long-Range Radar\\
			& (SRR208) & (ARS408)\\
			\hline
			5 meters & $\pm$60° & $\pm$60°\\
			\hline
			50 meters & $\pm$60° & $\pm$40°\\
			\hline
			60 meters & Out of View & $\pm$40°\\
			\hline
			150 meters & Out of View & $\pm$9°\\
			\hline
			250 meters & Out of View & $\pm$4°\\
			\hline
		\end{tabular}
	\end{center}
	\vspace{-2em}
\end{table}
}
{\renewcommand{\arraystretch}{1.2}
\begin{table}[h]
	\caption{Detection rate SCN01}
	\label{table:use_case_01}
	\centering
	\begin{center}
		\begin{tabular}{|c||c|c|c|c|}
			\hline
			Object reference & Distance & Azimuth & SRR & LRR\\
			& (meters) & (degrees) &  & \\
			\hline
			OBJ05 & 10 & 10° & Strong & Strong \\
			\hline
			OBJ06 & 12 & 45° & Weak & Failed \\
			\hline
			OBJ13 & 20 & 30° & Strong & Strong \\
			\hline
			OBJ02 & 30 & 0° & Strong & Strong \\
			\hline
			OBJ01 & 45 & 20° & Strong & Strong \\
			\hline
			OBJ03 & 40 & 5° & Strong & Strong \\
			\hline
		\end{tabular}
	\end{center}
\end{table}
}
{\renewcommand{\arraystretch}{1.2}
\begin{table}[h]
	\caption{Detection rate SCN02}
	\label{table:use_case_02}
	\begin{center}
		\begin{tabular}{|c||c|c|c|c|}
			\hline
			Object reference & Distance & Azimuth & SRR & LRR\\
			& (meters) & (degrees) &  & \\
			\hline
			OBJ05 & 10 & 20° & Strong & Strong \\
			\hline
			OBJ06 & 12 & -20° & Strong & Strong \\
			\hline
			OBJ13 & 20 & 10° & Weak & Strong \\
			\hline
			OBJ02 & 30 & -10° & Strong & Strong \\
			\hline
			OBJ03 & 40 & -5° & Strong & Strong \\
			\hline
			OBJ04 & 40 & 5° & Strong & Strong \\
			\hline
			OBJ12 & 35 & 0° & Weak & Strong \\
			\hline
		\end{tabular}
	\end{center}
\end{table}
}
{\renewcommand{\arraystretch}{1.2}
\begin{table}[h]
	\caption{Detection rate SCN03}
	\label{table:use_case_03}
	\begin{center}
		\begin{tabular}{|c||c|c|c|c|}
			\hline
			Object reference & Distance & Azimuth & SRR & LRR\\
			& (meters) & (degree) &  & \\
			\hline
			OBJ07 & 5 & 45° & Strong & Strong \\
			\hline
			OBJ11 & 20 & -15° & Strong & Strong \\
			\hline
			OBJ08 & 25 & 0° & Failed & Failed \\
			\hline
			OBJ09 & 40 & 10° & Strong & Strong \\
			\hline
			OBJ10 & 40 & -10° & Weak & Strong \\
			\hline
		\end{tabular}
	\end{center}
\end{table}
}

Tables~\ref{table:use_case_01},~\ref{table:use_case_02} and ~\ref{table:use_case_03} show the collected results from our tests. In these tables, objects are referenced by the label given in Figure~\ref{fig:objects_list}. The distance and the azimuth angle, which is counter-clockwise, is also included in order to know how this variable could affect the detection. The words used to notify the object detection by the radar are: \textit{strong}, \textit{weak}, and \textit{failed}. \textit{Strong} means the radar was able to detect the object all-time on the recorded data. On the other hand, \textit{weak} is used when the object was not detected the whole time during the test and the detection presents some flickering. Finally, \textit{failed} is used when there are no detection messages from the radar. 

According to our results, the LRR presents better performance than the SRR. However, both radars were able to detect most of the objects placed on each scenario. OBJ06 was barely detected by the SRR and missed by the LRR in the SCN01 but, by contrast, it was detected without problem in the SCN02. That means that the azimuth of the object was too wide and got out of the range of the radar. Another remark is the detection of the OBJ13, which is strong in the SCN02 but, presents problems when it is detected in the scenario two by SRR even when it would be easier to detect due to a narrower azimuth angle. A point to consider is the OBJ12 in the SCN02, which is detected by both radars even when the object is relatively far away from both radars and the object height is less than half a meter. 

From the results of SCN03, most of the objects were also detected. Only two objects have detection problems. OBJ08 was neither detected by the LRR nor the SRR. This is within our expectations since glass recognition is not optimal in radars and also the small size object difficults its recognition. OBJ10 was strongly detected by the LRR but not by the SRR, which could be caused by the shape of the scooter. Although it is completely metallic, the thin and small shape, together with the object distance, could compromise the recognition by the radar. 

\section{CONCLUSIONS}
\label{section:conclusions}

This paper describes a work-in-progress implementation in which only the first evaluations were carried out. These results show that, although both LRR and SRR radars present similar performance, the LRR was able to detect more objects from the defined scenarios.  

In-depth analysis of selected scenarios is planned in future work where not just the detection capabilities but the physics-based reasons of the ambiguous detection are investigated. For the purpose, we are planning (i) to design extended tests (e.g. in the case of OBJ08, different size bottles filled with different liquids in various relative positions are tested), (ii) to simulate the scattered electromagnetic waves and (iii) to consider the recognition algorithms of the radar. Based on the planned investigations we try to outline the possibilities of detection for the set of scenarios studied. In addition, by these systematic investigations we aim to ground a methodology to analyse situations when the radar sensor is found to function unreliably. By giving the reasons of the ambiguous operation we hope radar manufacturers can gain useful information to improve their designs too. From the integration perspective, collected sensor data will be fused to improve,  and further the development of functions included on the SPIDER such as the collision avoidance function.

\section*{ACKNOWLEDGMENT}

The publication was written in the project "LiDcAR". The project was funded by the program "Mobilität der Zukunft" of the Austrian Federal Ministry for Climate Action (BMK). The publication was written at Virtual Vehicle Research GmbH in Graz and partially funded within the COMET K2 Competence Centers for Excellent Technologies from the Austrian Federal Ministry for Climate Action (BMK), the Austrian Federal Ministry for Digital and Economic Affairs (BMDW), the Province of Styria (Dept.12) and the Styrian Business Promotion Agency (SFG). The Austrian Research Promotion Agency (FFG) has been authorised for the programme management. They would furthermore like to express their thanks to their supporting industrial and scientific project partners, namely Automotive Proving Ground Zala Ltd. and to the Budapest University of Technology and Economics. This work has been partially funded by EU ECSEL Project SECREDAS. This project has received funding from the European Unions Horizon 2020 research and innovation programme under grant agreement No 783119.

\bibliographystyle{IEEEtran}
\bibliography{bibliography}

\end{document}